\documentclass[conference]{IEEEtran}
\IEEEoverridecommandlockouts
\usepackage{cite}
\usepackage{amsmath,amssymb,amsfonts}
\usepackage{algorithmic}
\usepackage{graphicx}
\usepackage{textcomp}
\usepackage{xcolor}
\usepackage{url} 
\usepackage{subfigure}
\usepackage{multirow}
\usepackage{todonotes}
\def\BibTeX{{\rm B\kern-.05em{\sc i\kern-.025em b}\kern-.08em
    T\kern-.1667em\lower.7ex\hbox{E}\kern-.125emX}}
\begin{document}

\title{HelixMO: Sample-Efficient Molecular Optimization in Scene-Sensitive Latent Space
}


\author{
  Zhiyuan Chen* \\
  Baidu Inc. \\
  Shenzhen \\
  \texttt{chenzhiyuan05@baidu.com} \\
   \and
  Xiaomin Fang* \\
  Baidu Inc. \\
  Shenzhen \\
  \texttt{fangxiaomin01@baidu.com} \\
  \and
  Zixu Hua\\
  Baidu Inc. \\
  Shenzhen \\
  \texttt{huazixu@baidu.com} \\
  \and
  Yueyang Huang\\
  Baidu Inc. \\
  Shenzhen \\
  \texttt{huangyueyang@baidu.com} \\
  \and
  Fan Wang\\
  Baidu Inc. \\
  Shenzhen \\
  \texttt{wang.fan@baidu.com} \\
  \and
  Hua Wu\\
  Baidu Inc. \\
  Beijing \\
  \texttt{wu\_hua@baidu.com} \\
}

\maketitle

\begin{abstract}
Efficient exploration of the chemical space to search the candidate drugs that satisfy various constraints is a fundamental task of drug discovery. Advanced deep generative methods attempt to optimize the molecules in the compact latent space instead of the discrete original space, but the mapping between the original and latent spaces is always kept unchanged during the entire optimization process. The unchanged mapping makes those methods challenging to fast adapt to various optimization scenes and leads to the great demand for assessed molecules (samples) to provide optimization direction, which is a considerable expense for drug discovery. To this end, we design a sample-efficient molecular generative method, HelixMO, which explores the scene-sensitive latent space to promote sample efficiency. The scene-sensitive latent space focuses more on modeling the promising molecules by dynamically adjusting the space mapping by leveraging the correlations between the general and scene-specific characteristics during the optimization process. Extensive experiments demonstrate that HelixMO can achieve competitive performance with only a few assessed samples on four molecular optimization scenes. Ablation studies verify the positive impact of the scene-specific latent space, which is capable of identifying the critical characteristics of the promising molecules. We also deployed HelixMO on the website PaddleHelix (\url{https://paddlehelix.baidu.com/app/drug/drugdesign/forecast}) to provide drug design service.
\end{abstract}


\section{Introduction}

\begin{figure}
\includegraphics[width=0.5\textwidth]{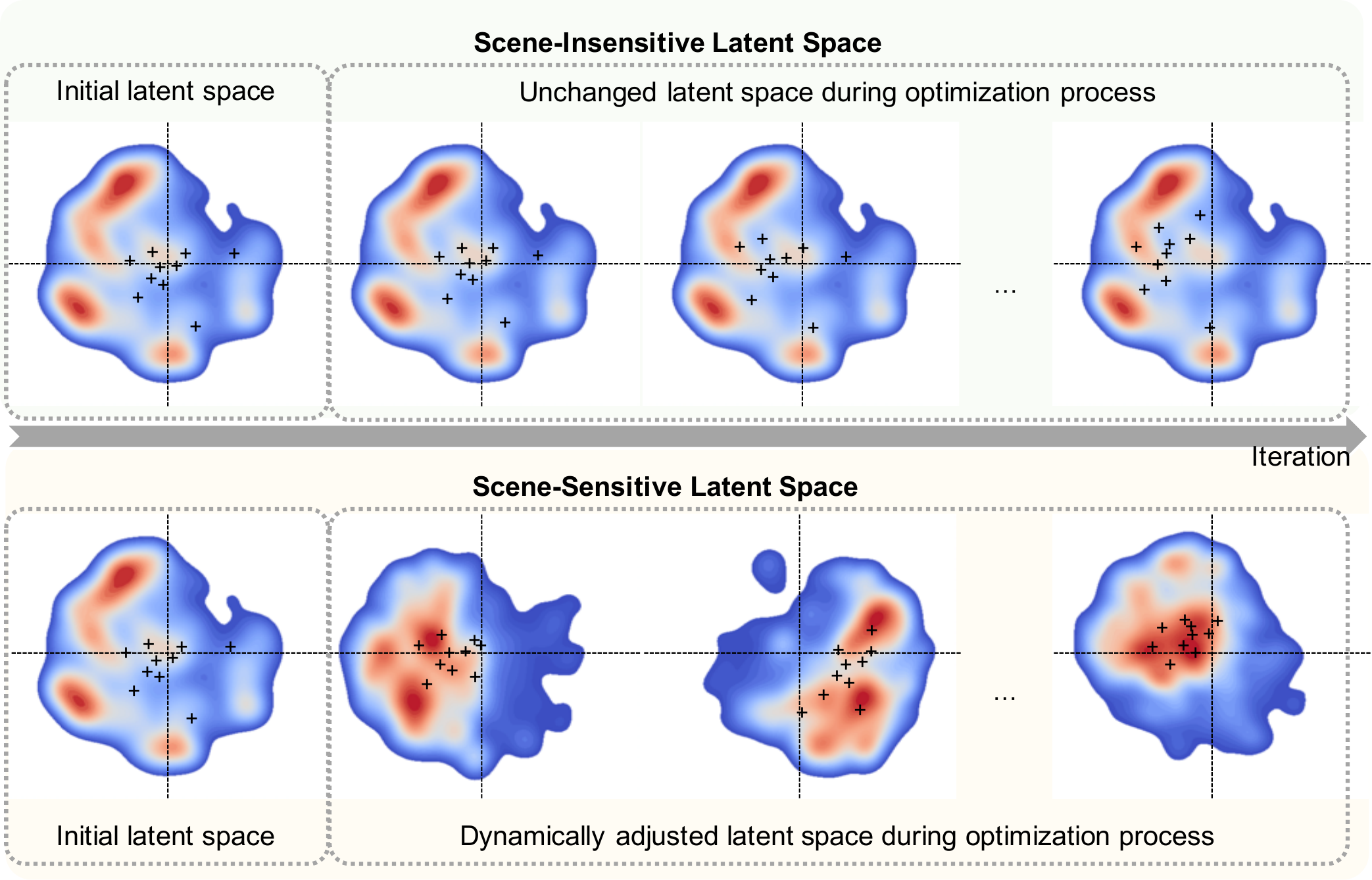}
\caption{Demonstration of the scene-insensitive latent space and scene-specific latent space. (Each heat map represents the distribution of molecules at a particular iteration. The redder the color in the region, the higher the properties of the molecules distributed here, and the bluer the color, the lower the properties of the molecules distributed there. The cross signs indicate the produced molecules of the generative method at each iteration.) When exploring the scene-insensitive latent space, the latent space is unchanged during the optimization process. In contrast, when exploring the scene-sensitive latent space, the latent space is adjusted during the optimization process, and the molecules with satisfactory properties tend to get together around the center of the space, focusing on modeling promising molecules.}
\label{fig:intro}
\end{figure}

Efficiently exploring the chemical space to search for molecules that meet various requirements (e.g., bio-activities, druggability, and synthetics accessibility) is one of the most critical tasks in the drug discovery industry. High-throughput screening (HTS) \cite{macarron2011impact} through laboratory experiments and visual screening (VS) \cite{bajorath2002integration,pagadala2017software} through in-silico experiments are usually involved in drug development. These approaches screen molecules from the molecular databases \cite{irwin2005zinc,gaulton2012chembl} to find promising molecules for further validation. However, the entire chemical space is at the scale of $10^{80}$ \cite{kirkpatrick2004chemical}, far exceeding the scale of known molecules in the databases. Consequently, in many scenarios, the databases used may not contain the satisfactory molecules that can be developed into drugs. Thus, developing more efficient and effective chemical space exploration methods is appealing for drug discovery.

Advanced studies proposed various fantastic deep molecular generative models, e.g., recurrent-based models \cite{segler2018generating,bjerrum2017molecular}, Variational AutoEncoder (VAE) \cite{doi:10.1021/acscentsci.7b00572,dai2018syntax,Jin2018,simonovsky2018graphvae,samanta2019nevae}, Generative Adversarial Network (GAN) \cite{yu2017seqgan,Guimaraes2016,decao2018molgan}, and Flow-based models \cite{madhawa2019graphnvp,honda2019graph,shi2020graphaf,Zang_2020}, to generate novel molecules that are not contained in the molecular libraries. They attempt to apply these deep molecular generative models to design de-novo drugs to produce promising molecules that meet various constraints and could be developed into drugs. At each iteration, the generative methods query the scoring system, e.g., laboratory experiments, to assess the multiple properties of some candidate molecules, e.g., measuring the bio-activities through the HTS, and expect the assessed molecules to provide the direction for molecular optimization. Particularly, some methods \cite{doi:10.1021/acscentsci.7b00572, dai2018syntax, Jin2018, Zang_2020} explore in the compact chemical latent space instead of the discrete original space (input space). The mapping between the input and latent spaces is learned from large-scale molecular libraries. Although such methods have demonstrated their effectiveness in searching the satisfactory molecules, they still 
need to further improve the sample efficiency to enhance the practicability in drug discovery because a large number of laboratory experiments for sample assessments would be a great expense for the drug R \& D institutions. We argue that designing a scene-sensitive latent space that can fast adapt to the optimization scenes during the exploration will likely benefit the sample efficiency. Previous latent-based works tend to keep the mapping of input and latent spaces unchanged during the whole exploration process. However, the distribution of the satisfactory molecules for a particular optimization scene is usually different from that of the molecules in the libraries used for learning the initial mapping. Consequently, it is struggled to explore the promising molecules in the latent space based on the distribution of the molecular libraries. The promising molecules for a particular optimization scene could be far away from the center of the latent space, and it may take a great many steps to explore the space in order to find them (as shown in the top of Fig.~\ref{fig:intro}). 

To this end, we design a sample-efficient molecular optimization method, HelixMO, exploring the scene-sensitive latent space to search the promising molecules. The mapping of the input and latent spaces is dynamically adjusted to adapt to the particular optimization scene. The space mapping is dynamically tuned by leveraging the assessed samples during the exploration process to focus on modeling the promising molecules gradually. Three kinds of learning tasks, including the reconstruction task, the scene-specific property prediction tasks, and the general property prediction tasks, are designed to encourage the latent space to infer the critical characteristics of the promising molecules. In such a way, the promising molecules will gradually gather together around the center of the latent space (as shown at the bottom of Fig.~\ref{fig:intro}), and only a few steps (samples) are required to find the satisfactory molecules.

Compared with several advanced baseline methods, HelixMO achieves competitive performance with fewer assessed samples on four molecular optimization scenes with one or more optimization objectives, indicating its potential values in practice. We also analyze the impacts of the learning tasks on modeling scene-sensitive latent space through ablation studies, verifying that HelixMO will pay more attention to the promising molecules to promote search efficiency. Furthermore, we have already deployed HelixMO by taking molecular docking as the scoring system on the website PaddleHelix\footnote{\url{https://paddlehelix.baidu.com/app/drug/drugdesign/forecast}} to provide drug design service.

Our main contributions can be summarized as follows:
\begin{itemize}
\item We designed a sample-efficient molecular optimization method, HelixMO, aiming to reduce the required number of assessed samples to improve the practicability in drug discovery.
\item The optimization method exploits the scene-sensitive latent space, which is dynamically adjusted by various learning tasks on limited assessed samples to adapt to the particular optimization scenes.
\item The proposed method can achieve competitive performance with fewer assessed samples than several baseline methods in four molecular optimization scenes, exhibiting the potential practical values.
\end{itemize}

\section{Preliminary}
\begin{figure*}[t]
\label{fig:framework}
\center
\includegraphics[width=0.9\textwidth]{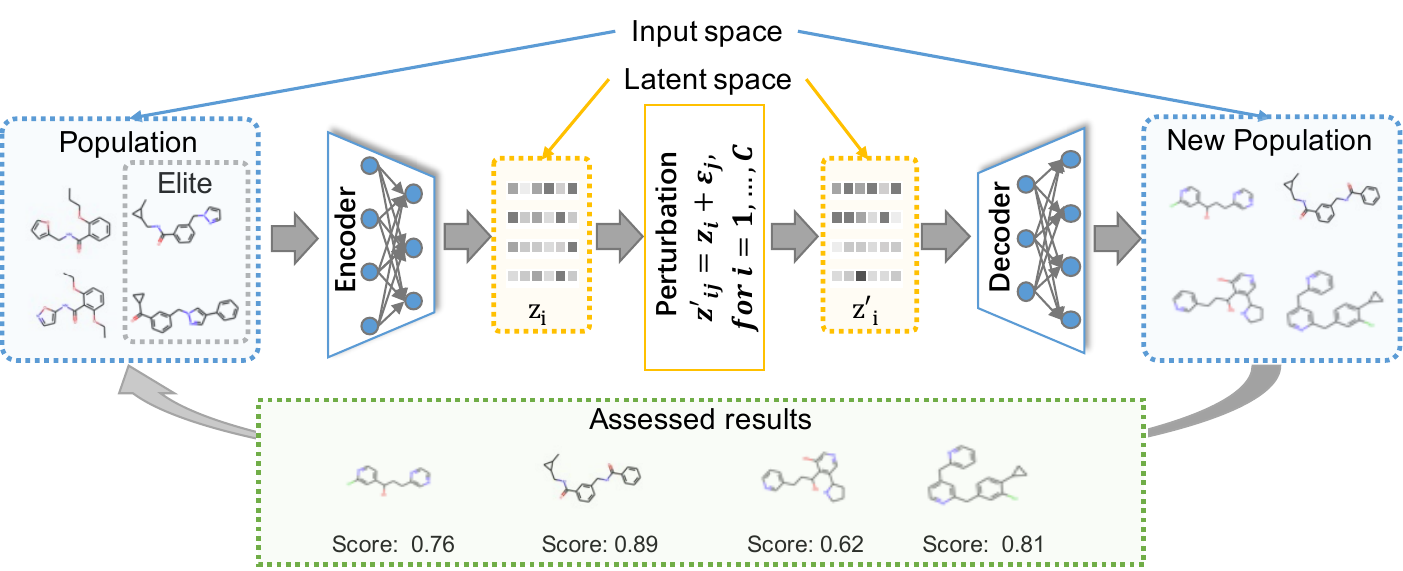}
\caption{Framework of a typical molecular optimization method that is based on exploration in latent space.}
\end{figure*}
Since HelixMO explores the latent space to search the molecules that satisfy multiple constraints, we first introduce a typical generative method based on latent space exploration in this section, as demonstrated in Fig.~\ref{fig:framework}. A genetic algorithm \cite{salimans2017evolution} is adopted as the search algorithm to optimize the molecules according to the assessed results from the scoring system. Usually, the molecular optimization process iterates multiple times until satisfactory molecules are found.

\subsection{Scoring System}
The scoring system assesses the molecules produced by the generative method at each iteration. An assessment of a molecule $x$ is formalized as
\begin{equation}
    A(x)=a_{1}(x) \circ a_{2}(x) \circ \cdots \circ a_{K}(x),
\end{equation}
which is composed of one or multiple properties, e.g., bio-activity to a target protein, druggability, and synthetic accessibility. $a_{k}(x)$ denotes the $k$-th objective, $K$ is the number of properties, and $\circ$ is an operator that combines the properties, such as summation and multiplication. When applying molecular optimization to drug development in practice, the properties of the molecules are evaluated by laboratory or in-silico experiments. Since those experiments are expensive and time-consuming, scoring functions \cite{gohlke2000knowledge} are usually used as alternatives for fast verification of the effectiveness of the molecular optimization methods. At the $i$-th iteration, the scoring system returns the overall assessed result $A(x)$ as well as all the properties, i.e., $a_{1}(x), \cdots$, and $a_{K}(x)$ for each molecule in the molecular population $x \in \mathcal{X}^{(i)}$ to the generative method, where $\mathcal{X}^{(i)}=\{x_1^{(i)}, x_2^{(i)}, \cdots, x_N^{(i)}\}$, and $N$ is the size of the population. 

\subsection{Generative Method}
We apply a genetic algorithm to search the satisfactory molecules. The properties of the molecular population produced at each interaction will gradually improve and meet the constraints of the optimization scenes.

At the $i$-th iteration, the generative method produces a new molecular population $\mathcal{X}^{(i)}$ by perturbing the molecules in the last population $\mathcal{X}^{(i-1)}$ according to the corresponding results assessed by the scoring system. Note that at the $1$-st iteration, the molecules in $\mathcal{X}^{(0)}$ are randomly selected from a drug-like molecular database.

First, each molecule $x \in \mathcal{X}^{(i)}$ is encoded into a representation vector through the encoder $z = Enc(x; \bf{\theta}^{Enc})$, where $\bf{\theta}^{Enc}$ denotes the parameters. Second, we select the elite molecules from the molecular population. The assessment score from the scoring system is formalized as a fitness score of a molecule. The representation vectors from $\mathcal{Z}^i$ are sampled and selected with a probability proportional to their fitness scores. Third, for each selected representation $z$, we randomly sample $C$ Gaussian noises $\epsilon_1, \epsilon_2, \cdots, \epsilon_c \sim N(0, \bf{I})$, where $C$ is a hyper-parameter to control the number of candidates and $\bf{I}$ represents an identity matrix. The Gaussian noises are added to $z$:
\begin{equation}
    z_{j}'=z+\sigma\cdot\epsilon_j,\enspace for \enspace j = 1, \cdots, C,
\end{equation}
where $z_{j}'$ is a perturbed representation vector, and $\sigma$ is the step size controlling the similarity degree between the $z$ and $z_{j}'$. If $\sigma$ is too large, there will be a great difference between the original molecule and the perturbed molecule, and the perturbed molecule may fail to retain the characteristics of the original molecule. If $\sigma$ is too small, the perturbed molecule will be particularly similar to the candidate molecule, resulting in lower efficiency of chemical space exploration. Fourth, the perturbed representations are reconstructed into the perturbed molecules by decoder $Dec(z; \bf{\theta}^{Dec})$, where $\bf{\theta}^{Dec}$ denotes the parameters.


Finally, the scoring system is queried to assess the properties of the molecules $\mathcal{X}^{(i)}$, and the set of assessed results is denoted as $\mathcal{A}^{(i)} =\{A(x_1^{(i)}), A(x_2^{(i)}), \cdots, A(x_N^{(i)})\}$ with $A(x)$ representing the assessed result of molecule $x$.

\section{Scene-Sensitive Latent Space}
\begin{figure*}[!t]
\center
\includegraphics[width=0.75\textwidth]{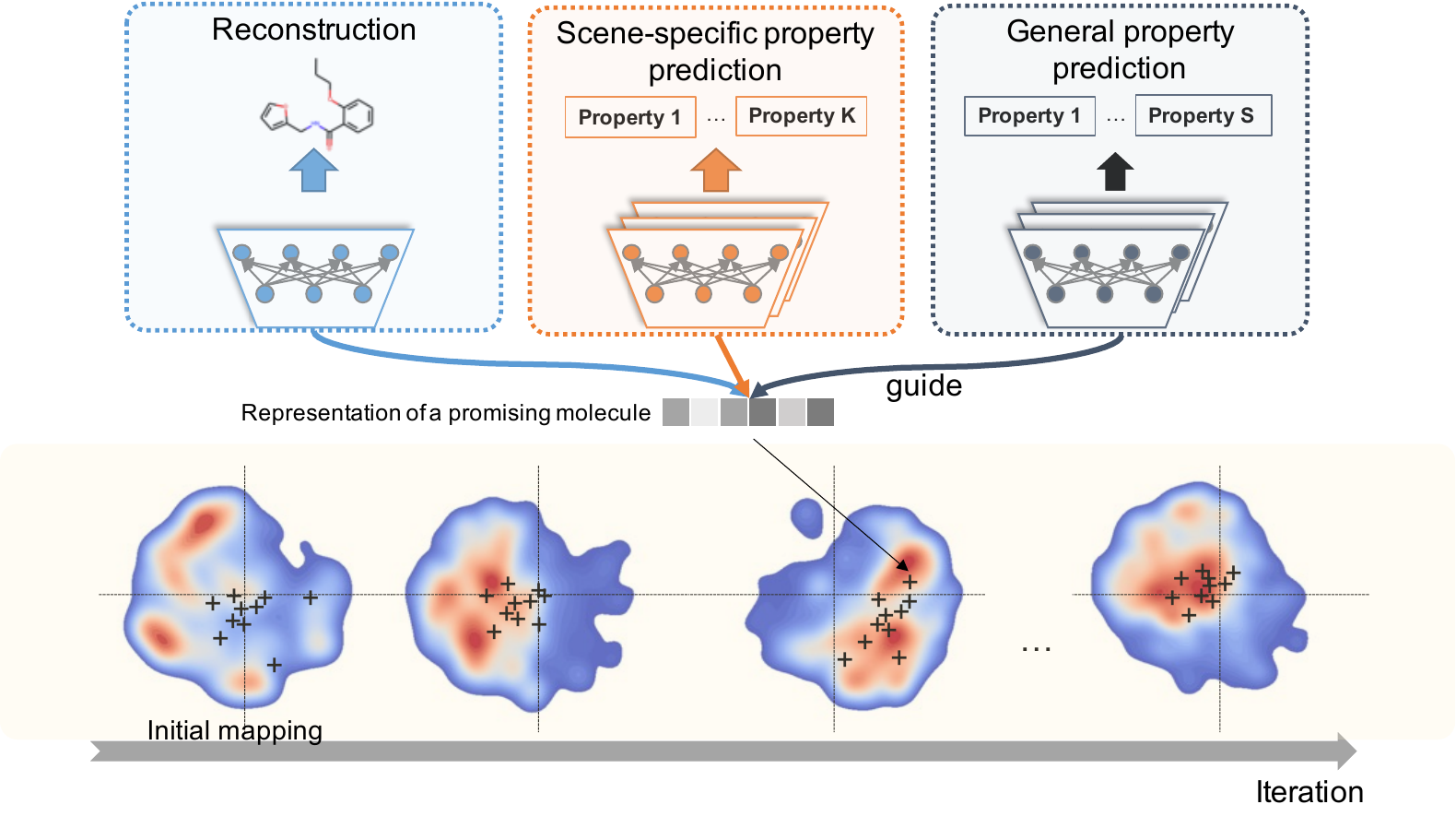}
\caption{HelixMO gradually adjusts the mapping of the input and latent spaces to focus on modeling the more promising molecules so as to fast adapt the particular optimization scene. Three kinds of learning tasks are leveraged to guide the adjustment of the latent space.}
\label{fig:scene_sensitive}
\end{figure*}

HelixMO searches the promising molecules in the scene-sensitive latent space. Since the distribution of molecules in the drug-like database used to obtain the initial mapping of the spaces is usually different from that of promising molecules of a particular optimization scene, we adjust the mapping of the input and latent spaces to encourage the latent space to focus more on modeling the promising molecules. The molecules with good properties tend to get together around the center of the latent space during the optimization process, and thus we can find the molecules that can satisfy multiple constraints more sample efficiently. The architecture of HelixMO is exhibited in Fig.~\ref{fig:scene_sensitive}. The mapping is dynamically adjusted in accordance with the potential of molecules discovered through three kinds of learning tasks, including the reconstruction task, scene-specific property prediction tasks, and general property prediction tasks, during the optimization process. Those learning tasks encourage the generative model to identify the promising molecules' essential characteristics, e.g., the task-specific properties and the general properties. Those essential characteristics can effectively distinguish the high-potential molecules from the remaining ones.

In this section, we first briefly introduce the initial mapping of the input and latent spaces. Then, we describe three kinds of learning tasks used to guide the adjustment of the mapping one after another.

\subsection{Initial Mapping of Input and Latent Spaces}
To obtain the initial mapping between the input space $\mathcal{X}$ and the latent space $\mathcal{Z}$, we adopt a widely used molecular generative model, Junction Tree Variational Autoencoder (JT-VAE) \cite{Jin2018}, as the backbone generative model. JT-VAE is trained on a large-scale molecular database, e.g., ZINC \cite{doi:10.1021/acs.jcim.0c00675} and ChEMBL \cite{gaulton2017chembl}. MOSES \cite{Polykovskiy2018}, containing about 250K molecules extracted from the ZINC database \cite{doi:10.1021/acs.jcim.0c00675} is utilized to obtain the initial mapping of the input and latent spaces. The encoder in JT-VAE can encode a molecule $x$ into a representation vector $z$: $z=Enc(x; \bf{\theta}^{Enc})$, and the decoder in JT-VAE can reconstruct that molecule from the representation vector $x=Dec(z;\bf{\theta}^{Dec})$. In such a method, we can obtain the initial mapping of the input and latent spaces. 

\subsection{Latent Space Adjustment}
Previous works tend to keep the mapping unchanged during the whole optimization process. HelixMO gradually tunes the mapping during the optimization process to make it sensitive to particular optimization scenes. 

To encourage the latent space to focus more on the modeling of the satisfactory molecules, i.e., the molecules with better properties, HelixMO takes advantage of the molecules produced at each iteration $\mathcal{X}^{(i)}$, and the parameters of the generative model are optimized with those produced molecules at each iteration. Since the properties of the molecules produced at each iteration are improved by the genetic algorithm, the generative model will gradually focus on the molecules with extraordinary properties.

Three kinds of learning tasks, including the reconstruction task, the scene-specific property prediction tasks, and the general property prediction tasks, guide the mapping adjustment by optimizing the parameters of the generative model $\bf{\Theta}=\{\bf{\theta}^{Enc}, \bf{\theta}^{Dec}, \bf{\theta}_k^{Spec}, \bf{\theta}^{Gene}\}$. The overall loss function for model optimization is defined as:
\begin{equation}
    L^{Total}(\mathcal{X}) = L^{Reco}(\mathcal{X}) + \frac{1}{K} \sum_{k=1}^{K}L_k^{Spec}(\mathcal{X}) + \frac{1}{S}\sum_{s=1}^{S}L_s^{Gene}(\mathcal{X}),
\end{equation}
where $\mathcal{X}=\mathcal{X}^{(i)}$ at the $i$-th iteration. $L^{Reco}(\mathcal{X})$, $L_k^{Spec}(\mathcal{X})$, $L_S^{Gene}(\mathcal{X})$ denote the loss function of the reconstruction task, a scene-specific property prediction task, and a general property prediction task, respectively, where the detail definition will be introduced in the following subsections. $K$ is the number of the scene-specific properties used in the scene-specific property prediction tasks, and $S$ is the number of the general properties used in the general property prediction tasks. Those learning tasks attempt to mine knowledge from the limited samples assessed by the scoring system at each iteration. The knowledge is incorporated into the generative model to learn the scene-sensitive mapping between latent and input spaces. 

\subsection{Reconstruction Task}
The reconstruction task is originally used to train the generative model, i.e., JT-VAE, to obtain the initial mapping of the input and latent spaces, which reconstructs the molecules from the latent representations. We also adopt the reconstruction task to tune the adjustment of the mapping of the input and latent spaces. The reconstruction task aims to minimize the the distance between a input molecule $x \in \mathcal{X}$ and the reconstructed molecule $\hat{x}=Dec(Enc(x;\bf{\theta}^{Enc});\bf{\theta}^{Dec})$:
\begin{equation}
    L^{Reco}(\mathcal{X})=\sum_{x \in \mathcal{X}}L^{Dis}(x, \hat{x}),
\end{equation}
where $\mathcal{X}=\mathcal{X}^{(i)}$ at the $i$-th iteration, and $L^{Dis}(.)$ is a function measuring the distance between two molecules introduced in \cite{Jin2018}. Using the reconstruction task to train on the produced molecules at each iteration can effectively lift the probability of the reconstruction of the promising molecules.

\subsection{Scene-Specific Property Prediction Tasks}
In order to guide the latent space to distinguish the molecules with superior properties from the remaining ones for the optimization scene, we introduce the scene-specific property prediction tasks. Those tasks attempt to estimate the properties, i.e., $a_1(x), a_2(x), \cdots, a_K(x)$. For the $k$-th property, a Multilayer Perceptron (MLP), denoted by $MLP_k^{Spec}()$, takes the latent representation vector $z$ of a molecule $x$ as the input and outputs the estimated property:
\begin{equation}
    \hat{a}_k(x) = MLP_k^{Spec}(Enc(x;\bf{\theta}^{Enc});\bf{\theta}_k^{Spec}),
\end{equation}
where $\bf{\theta}_k^{Spec}$ denotes the parameters of of function $MLP_k^{Spec}()$. The parameters $\bf{\theta}^{Enc}$ and $\bf{\theta}_k^{Spec}$ are optimized by minimizing the distance between the estimated property $\hat{a}_i(x)$ and the property return from the scoring system $a_i(x)$ for a molecule $x$. At the $i$-th iteration, the assessed molecules collected from the scoring systems, i.e., $\mathcal{X}^{(i)}$, are utilized to optimize the parameters $\bf{\theta}^{Enc}$ and $\bf{\theta}_k^{Spec}$:
\begin{equation}
    L_k^{Spec}(\mathcal{X})=\sum_{x\in \mathcal{X}}Huber\_loss_\delta(\widetilde{a}_k(x),\hat{a}_k(x)),
    \label{eq:loss_spec}
\end{equation}
where $\mathcal{X}=\mathcal{X}^{(i)}$ at the $i$-th iteration, and $Huber\_loss_{\delta}(.)$ \cite{huber1992robust} is a loss function that measures the distance of two values with $\delta$ controlling the impact of the outliers. As the number of iterations increases, the generative method would produce more satisfactory molecules, and the property values of the assessed molecules would become larger. To alleviate the negative impact caused by the changing scale of property values for model training, we normalize the property values by applying Gaussian Normalization, and the normalized property values are denoted as $\widetilde{a}_k(x)$ in Eq.~\ref{eq:loss_spec}. Through the scene-specific property prediction tasks, the model can infer the scene-specific properties of a molecule from its latent representation vector.

Note that the scene-specific property prediction tasks can also serve as a coarse scoring system to screen the molecules with more potential. HelixMO uses this coarse scoring system to screen out the perturbed molecules with the highest scores assessed by the coarse scoring system as the molecular population of the next iteration to improve the sample efficiency further.

\subsection{General Property Prediction Tasks}
We also introduce the general property prediction tasks to further enhance the capacity of the latent representations. Molecular fingerprints \cite{rogers2010extended} can encode a molecule into a list of binary bits, describing the presence of the structure fragments in the molecule. The structure fragments of a molecule are strongly correlated to its general properties. Consequently, each binary bit can be simply regarded as a general property of a molecule. For the sake of identifying the assessed molecules' general properties, the general property prediction tasks simultaneously estimate the binary bits of the molecular fingerprints from their latent representations:
\begin{equation}
    \hat{b}_1(x), \cdots, \hat{b}_S(x) = MLP^{Gene}(Enc(x;\bf{\theta}^{Enc});\bf{\theta}^{Gene}),
\end{equation}
where $\hat{b}_s(x)$ denotes the $s$-th estimated bit, and $\theta^{Gene}$ represents the parameters of the general property prediction tasks, denoted by function $MLP^{Gene}$. Cross-entropy is taken as the loss function to optimize the parameters $\bf{\theta}^{Enc}$ and $\bf{\theta}^{Gene}$ on the assessed molecules at the $i$-th iteration:
\begin{equation}
    L_s^{Gene}(\mathcal{X})=\sum_{x\in \mathcal{X}}Cross\_entropy(b_s(x), \hat{b}_s(x)),
\end{equation}
where $s$ is the index of the binary bits, and $\mathcal{X}=\mathcal{X}^i$. We utilize the Morgan fingerprint \cite{morgan1965generation} with 2048 bits as the general properties of the molecules. Learning the latent representations in this way is equivalent to using the self-supervised learning (SSL) method \cite{fang2022geometry} to learn valuable information from the unlabeled data, which has achieved great success in many domains.

\section{Experiments}
\subsection{Experimental Settings}
Following the previous work \cite{fu2022differentiable}, the molecular optimization methods are compared through four molecular properties:
\begin{itemize}
  \item GSK3$\beta$ \cite{li2018multi}: Inhibition against glycogen synthase kinase-3$\beta$;
  \item JNK3 \cite{li2018multi}: Inhibition against c-Jun N-terminal kinase-3;
  \item QED \cite{bickerton2012quantifying}: Quantitative Estimation of Drug-likeness;
  \item SA \cite{ertl2009estimation}: Synthetics Accessibility, indicating the difficulty of synthesizing a given molecule.
\end{itemize}
Each property is obtained through a scoring function. The properties compose four optimization scenes: two single-objective optimization scenes, including \emph{GSK3$\beta$} and \emph{JNK3}, and two multi-objective optimization scenes, including \emph{GSK3$\beta$+JNK3} and \emph{GSK3$\beta$+JNK3+QED+SA}. We use summation as the operator $\circ$ to combine the properties in multi-objective optimization scenes.

\subsection{Overall Evaluation}
\begin{figure*}[!t]
\center
\includegraphics[width=1.0\textwidth]{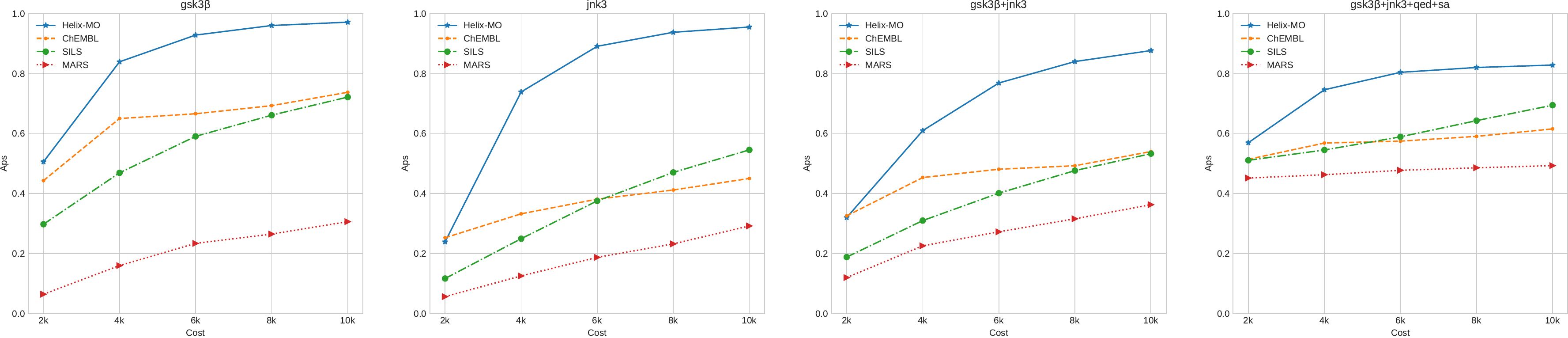}
\caption{Comparison of the sample efficiency of various methods with only a few assessed samples provided for molecular optimization on four optimization scenes.}
\label{fig:novar}
\end{figure*}


To verify that HelixMO has a great advantage in sample efficiency, we evaluate the performance of HelixMO and several typical baseline methods when the number of requests for the scoring system / the number of assessed samples (\emph{Cost} for molecular optimization) is set to be a specific number. We compare HelixMO with Molecular Database (ChEMBL) \cite{gaulton2017chembl}, MARS \cite{xie2021mars}, and Scene-Insensitive Latent Space (SILS). ChEMBL is a database of bioactive molecules with drug-like properties, which is used as a naive baseline based on virtual screening in our paper. We randomly sampled some molecules from the molecular database, and the top sampled molecules were evaluated. MARS explores the input space for molecular optimization, modeling the probability of adding/deleting fragments to/from the original molecules to produce new molecules. SILS takes JT-VAE \cite{Jin2018} as the model backbone and explores the unchanged latent space for molecular optimization by a genetic algorithm, just like HelixMO. ILS is a scene-insensitive ablation version of HelixMO. We use the default hyper-parameters that are recommended in the corresponding literature. For each method, the average property scores (\emph{Aps}s) of the top-100 molecules under different \emph{Cost}s (2K, 4K, 6K, 8K, and 10K) are reported in Fig.~\ref{fig:novar}. 
We operated five runs on HelixMO for each optimization scene and reported its average performance.

From the figure, we can draw the following conclusions:
\begin{itemize}
    \item Compared with the baseline molecular optimization methods, HelixMO demonstrates its effectiveness and efficiency in finding satisfactory molecules on four optimization scenes. The Aps scores of HelixMO significantly surpass those of the baseline methods in almost all the cases.
    \item Although SILS is also based on latent space exploration, HelixMO still shows superior sample efficiency. The dynamically adjusted latent space used by HelixMO can effectively identify the promising molecules by analyzing the information of molecules produced at each iteration, reducing the number of search steps required.
    \item Interestingly, although ChEMBL only randomly samples molecules from the database without reliance on the deep generative model, the Aps scores of ChEMBL are higher than those of MARS and SILS when only 2K assessed samples are available on four optimization scenes. We speculate that in the early stage of molecular optimization, the properties of the molecules in the regions explored by those two methods are unsatisfactory and even inferior to the drug-like molecules in ChEMBL.
\end{itemize}

\begin{figure*}[ht]
\centering    
\subfigure[Cost]{\label{fig:cost}\includegraphics[width=0.19\textwidth]{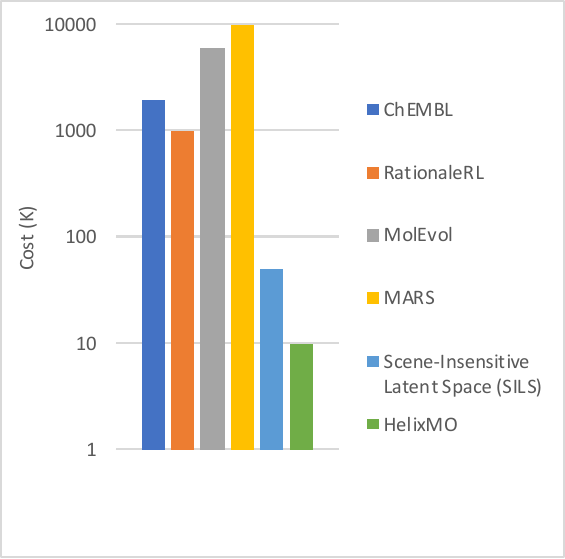}}
\quad
\subfigure[Aps]{\label{fig:aps}\includegraphics[width=0.37\textwidth]{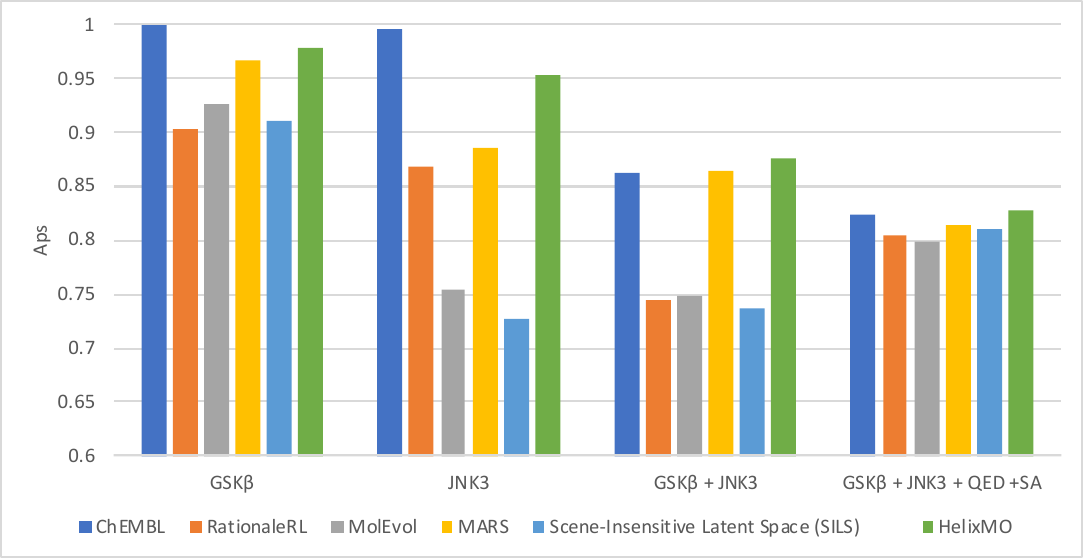}}
\quad
\subfigure[Nov]{\label{fig:nov}\includegraphics[width=0.37\textwidth]{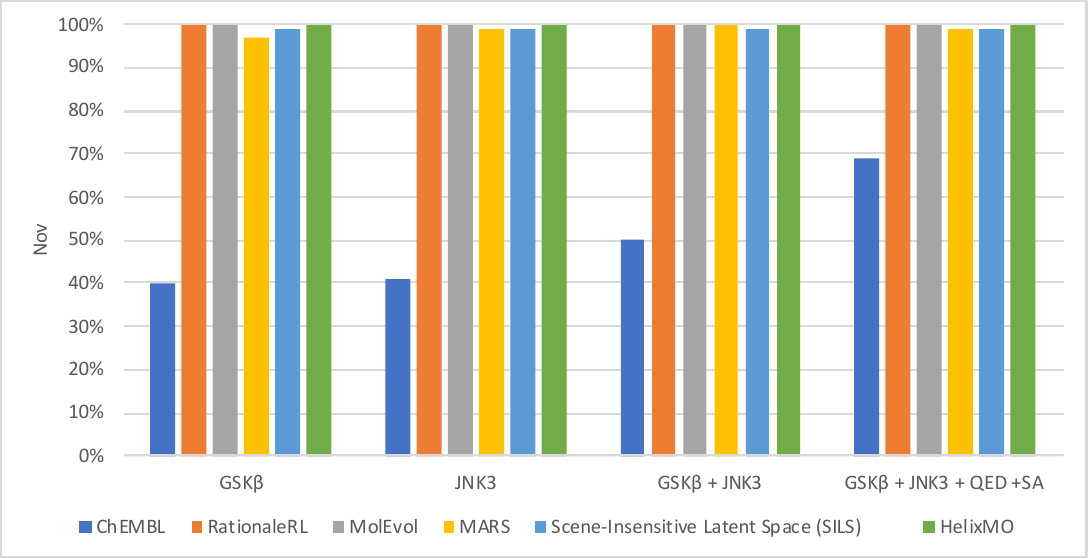}}
\caption{Overall evaluation of HelixMO and the baseline methods on four optimization scenes.}
\label{fig:overall_evaluation}
\end{figure*}

To comprehensively compare HelixMO with the advanced molecular generative methods, we evaluate HelixMO and more advanced baseline methods (RationaleRL \cite{Jin2020a} and MolEvol \cite{chen2020molecule}) from three perspectives: (1) \emph{Cost}: assessed samples required for molecular optimization; (2) \emph{Aps}: average property score of the top-100 molecules; and (3) Novelty (\emph{Nov}): the ratio of generated molecules not present in the known active molecule set toward targets GSK3$\beta$ and JNK3. The overall evaluation of HelixMO and the baseline methods on four optimization scenes are exhibited in Fig~\ref{fig:overall_evaluation}.

(1) As exhibited in Fig.~\ref{fig:cost}, some baseline methods, including RationaleRL, MolEvol, and MARS, require millions of assessed samples to search the satisfactory molecules. In contrast, the methods that explore the latent space, including SILS and HelixMO, require only tens of thousands of assessed molecules for chemical space exploration. Especially by applying the scene-sensitive latent space, only 10K assessed samples are required to achieve competitive performance with the baseline methods. These results demonstrate the sample efficiency of HelixMO.
(2) As exhibited in Fig.~\ref{fig:aps}, the \emph{Aps} scores of HelixMO are comparable to those of the optimal baseline methods on all the optimization scenes in general. For optimization scenes \emph{GSK3$\beta$} and \emph{JNK3} with only one optimization objective, HelixMO surpasses all the baseline methods, except ChEMBL. For the more difficult scenes with multiple optimization objectives, HelixMO achieves the highest \emph{Aps} scores on \emph{GSK3$\beta$+JNK3} and \emph{GSK3$\beta$+JNK3+QED+SA}. These results indicate that HelixMO is capable of discovering those molecules with satisfactory properties.
(3) Although the Aps scores of the top molecules in the drug-like database ChEMBL are excellent on four optimization scenes, as shown in Fig.~\ref{fig:nov}, the Nov scores of the top molecules are the worst among all the methods. Many top molecules in ChEMBL have already been contained in the known active molecule set toward targets GSK3$\beta$ and JNK3 since a virtual screening method cannot design de novo molecules that are not in the screening database.


\subsection{Ablation Studies}

\begin{figure*}[ht]
\centering    
\subfigure[Aps scores of ablation versions of HelixMO. Each learning tasks used to learn the scene-sensitive latent space contributes to improve Aps scores.]{\label{fig:ablation_prop4}\includegraphics[width=0.32\textwidth]{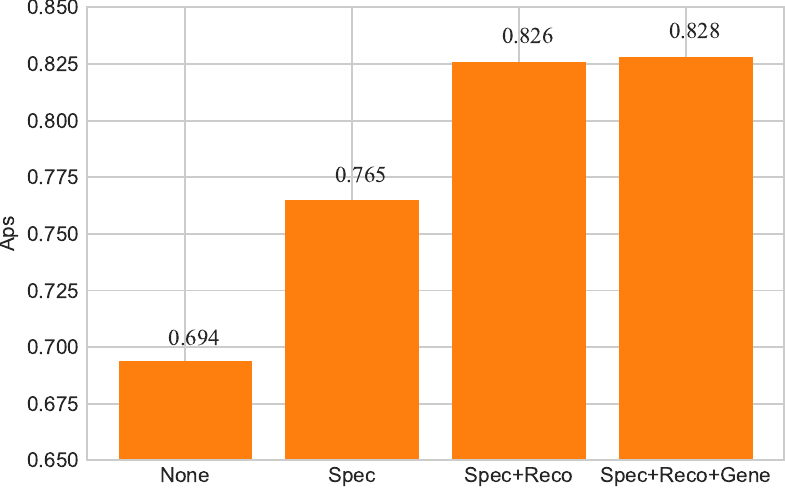}}
\quad
\subfigure[Average norms of the promising molecules' representation vectors produced by various ablation versions.]{\label{fig:ablation_center}\includegraphics[width=0.3\textwidth]{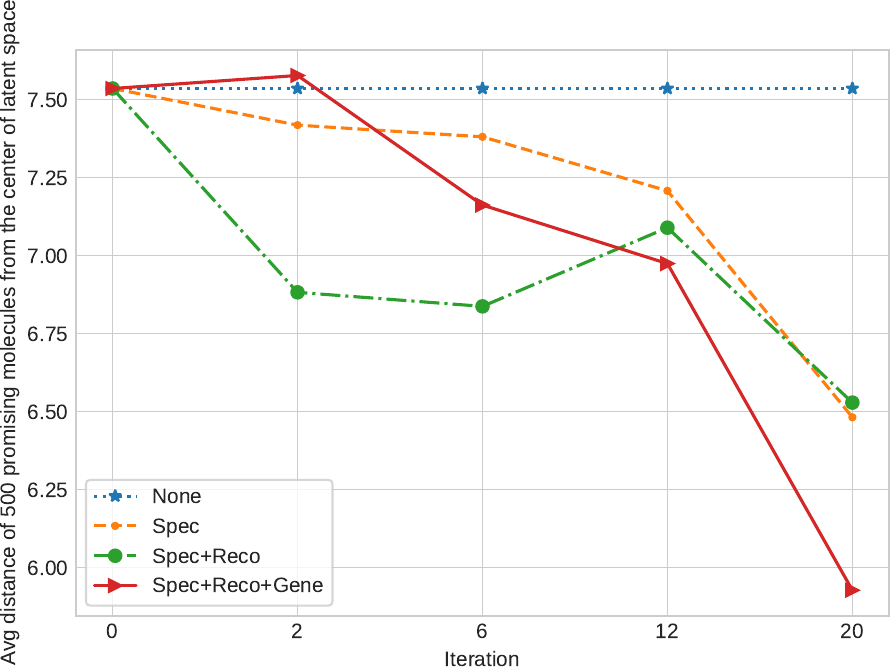}}
\quad
\subfigure[The average Aps scores as well as the standard deviation of the Aps scores of five independent runs of \emph{Spec+Reco+Gene} and \emph{Spec+Reco}. The average standard deviation of each method is shown in the  brackets.]{\label{fig:all_var}\includegraphics[width=0.3\textwidth]{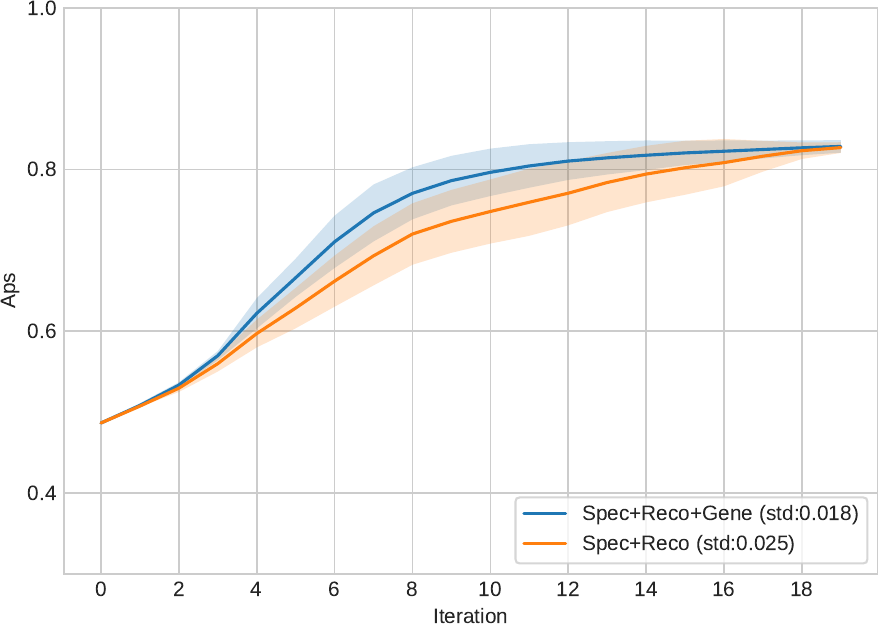}}
\caption{Ablation studies on optimization scene \emph{GSK3$\beta$+JNK3+QED+SA}.}
\end{figure*}

HelixMO makes efforts to fine-tune the mapping between the input space and latent space through three kinds of learning tasks for the sake of making the latent space sensitive to the optimization scenes. To explore the impacts of the learning tasks on the latent space, we compare multiple ablation versions of HelixMO with different settings:
\begin{itemize}
  \item \emph{None}: The latent space is kept unchanged during the molecular exploration process (refer to SILS in the last subsection);
  \item \emph{Spec}: The scene-specific property prediction tasks are used to guide the adjustment of the latent space during the molecular exploration process;
  \item \emph{Spec+Reco}: On the basis of \emph{Spec}, the reconstruction task is also utilized to fine-tune the model during the molecular exploration process.
  \item \emph{Spec+Reco+Gene}: The complete version of HelixMO with the reconstruction task, scene-specific property prediction tasks, and general property prediction tasks.
\end{itemize}
The hyper-parameters are kept the same for all the ablation versions of HelixMO, and only 10K assessed samples are available for molecular optimization.

Fig.~\ref{fig:ablation_prop4} exhibits the Aps scores of ablation versions of HelixMO on optimization scene \emph{GSK3$\beta$+JNK3+QED+SA}. (1) \emph{Spec} achieves better Aps scores than \emph{None}. The scene-specific property prediction tasks in \emph{Spec} encourage the model to infer the properties of the molecules from the latent space, and thus the model is capable of distinguishing the satisfactory molecules from the others. Besides, the scene-specific property prediction tasks also serve as a coarse scoring system to provide direction to choosing the candidate molecules with more potential. (2) The significant improvement of \emph{Spec+Reco} compared with \emph{Spec} indicates that the reconstruction task contributes to the improvement during the molecular exploration process. It is difficult for the generative model to sample/reconstruct a molecule that is out of the distribution of the molecules in a database used to obtain the initial mapping. Since the distribution of the satisfactory molecules for a particular optimization scene is usually different from that of the molecules in the database, fine-tune the latent space precipitate to enhance the success rate of reconstructing a satisfactory molecule. (3) By comparing \emph{Spec+Reco} and \emph{Spec+Reco+Gene}, we observe that the general property prediction tasks only bring a little improvement in the Aps scores. Although the general property prediction tasks can not explicitly promote the Aps scores, it still plays roles in better space mapping and enhancement of the robustness for molecular optimization, which will be further discussed in the following. 

The scene-sensitive latent space is the primary factor for the high sample efficiency of HelixMO. A scene-sensitive mapping between the input and latent space should focus more on modeling the promising molecules for the particular optimization scenes rather than molecules in the drug-like database. The VAE-style backbone model used by HelixMO assumes the molecules follow the normal distribution in the latent space. Therefore, the attention degree of the model to a molecule is inversely proportional to the distance of the molecule from the center of the latent space (the norm of the representation vector). We believe that the closer the promising molecules are to the center of the latent space, the faster the genetic algorithm will find these molecules that meet the property constraints. Fig.~\ref{fig:ablation_center} shows the average norms of the representation vectors in various latent spaces of 500 promising molecules for optimization scene \emph{GSK3$\beta$+JNK3+QED+SA}. More concretely, we calculated the average norm of the encoded representation vectors of the promising molecules at multiple iterations for each ablation version of HelixMO. For \emph{Spec}, \emph{Spec+Reco}, and \emph{Spec+Reco+Gene}, the average norms decrease with the increase of iteration, which means the dynamically adjusted latent spaces of these methods gradually pay more attention to the promising molecules. Significantly, the average norm for \emph{Spec+Reco+Gene} at the last iteration is the lowest, indicating that the general property prediction tasks can strengthen the modeling ability for the molecules that meet the requirements of the optimization scenes.

Besides, the general property prediction tasks can be regarded as auxiliary self-supervised learning tasks, which can improve the robustness of the generative model. Fig.~\ref{fig:all_var} shows the average Aps scores as well as the standard deviation of the Aps scores of five independent runs of \emph{Spec+Rec+Gene} and \emph{Spec+Reco} at multiple iterations on optimization scene \emph{GSK3$\beta$+JNK3+QED+SA}. \emph{Spec+Rec+Gene} with the general property prediction tasks can not only find satisfactory molecules with fewer assessed samples but also has better robustness, where the deviation of the Aps is smaller than that of \emph{Spec+Reco} during the molecular exploration process.

\section{Conclusions}
The development of deep generative models makes it possible to search the entire chemical space to find potential drugs. Due to the low sample efficiency, employing the existing generative methods in practical applications is still challenging. In order to reduce the number of the required assessed samples, we developed HelixMO, attempting to explore the scene-sensitive latent space to efficiently search the satisfactory molecules. The mapping of the input and latent space is dynamically adjusted to focus on modeling the promising molecules by three well-designed learning tasks. In this way, HelixMO is capable of achieving competitive performance with fewer assessed samples than the mainstream generative methods. We also deploy HelixMO as an online service, expecting to provide experimental inspiration for drug R \& D researchers.

\begin{appendices}
\section{Appendix: Model Architecture of Helix-MO}
\begin{figure*}[!t]
    \center
    \includegraphics[width=0.8\textwidth]{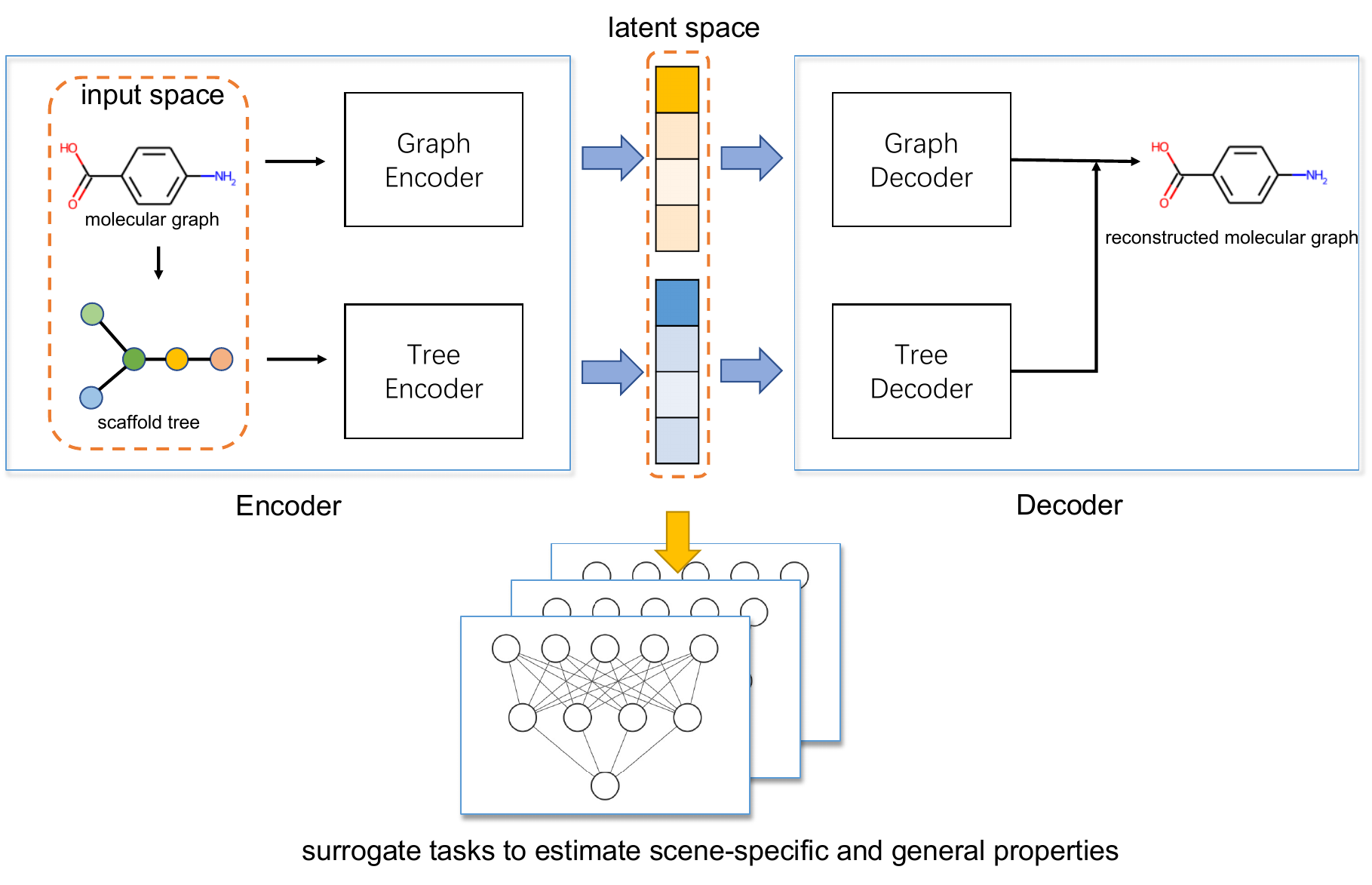}
    \caption{Model architecture of Helix-MO.}
    \label{fig:architecture}
\end{figure*}
We take JT-VAE as the basic model architecture and additionally add multiple heads on the latent representation to estimate the scene-specific and general properties. The model architecture of Helix-MO is shown in Fig.~\ref{fig:architecture}. The input of the generative model is a molecular graph \emph{G} and a scaffold tree \emph{T}. The scaffold tree is generated from the molecular graph \emph{G} by tree decomposition, and each node represents a chemical fragment. Graph VAE and tree VAE are the main components of the model, consisting of an encoder and a decoder, respectively. The graph encoder is a three-layer Graph Neural Network (GNN), and the tree encoder is a twenty-layer GNN. The tree encoder propagates the messages from the root node to all the leaf nodes in the top-down phase, and then a GRU is adopted to send and update the messages in the tree. For each encoder, the dimension of the hidden layer is 450, and the output dimension is 28. The graph decoder uses a message-passing network to derive a candidate subgraph vector representation. The tree decoder uses GRU to decode the scaffold tree in a depth-first traversal order. Multiple multi-layer perceptions (MLPs) are exploited as the model heads to predict the properties. Specifically, for each property (including scene-specific and general properties), a three-layer MLP with a hidden size of 56 is used to take the latent vector as input to predict the molecular property score.

\section{Appendix: Experimental Settings\label{appendix:experimental_settings}}

\subsection{Evaluation Metrics\label{appendix:experimental_settings_metrics}}
\begin{itemize}
  \item GSK3$\beta$ \cite{li2018multi}: Inhibition against glycogen synthase kinase-3$\beta$. The score of GSK3$\beta$ for a molecule is evaluated by a Random Forest, which is trained on the dataset with 2 thousand positive samples and 50 thousand negative samples. 
  \item JNK3 \cite{li2018multi}: Inhibition against c-Jun N-terminal kinase-3. The score of JNK3 for a molecule is evaluated by a Random Forest trained on the dataset with 7 hundred positive samples and 50 thousand negative samples.
  \item QED \cite{bickerton2012quantifying}: Quantitative Estimation of Drug-likeness. QED is an integrative score to evaluate molecules’ favorability to become a drug.
  \item SA \cite{ertl2009estimation}: Synthetics Accessibility. SA indicates the difficulty of synthesizing a given molecule. The raw SA scores range from 1 to 10. The higher the SA score is, the more complex the molecule is to be synthesized. We normalize the SA scores into range $[0.0, 1.0]$ \cite{doi:10.1021/acs.jcim.0c00174}:     
  \begin{equation}
  \text { normalized-SA }(x)= \begin{cases}1, & \mathrm{SA}(x)<\mu; \\ \exp \left(-\frac{(\mathrm{SA}(x)-\mu)^{2}}{2 \sigma^{2}}\right), & \mathrm{SA}(x) \geq \mu,\end{cases}      
  \end{equation}
  where $\mu = 2.230044$, $\sigma = 0.6526308$.
\end{itemize}

\subsection{Practicability of HelixMO for Drug Discovery}
We attempted to investigate whether HelixMO can effectively discover the candidate drugs in practice. HelixMO is already deployed in the website \url{https://paddlehelix.baidu.com/app/drug/drugdesign/forecast} to provide a drug discovery service. A widely used docking tool AutoDock Vina \cite{trott2010autodock} and a AI-based model, KDeep \cite{doi:10.1021/acs.jcim.7b00650} are used as the external system to provide assessments.

\begin{figure*}[!t]
\center
\includegraphics[width=1.0\textwidth]{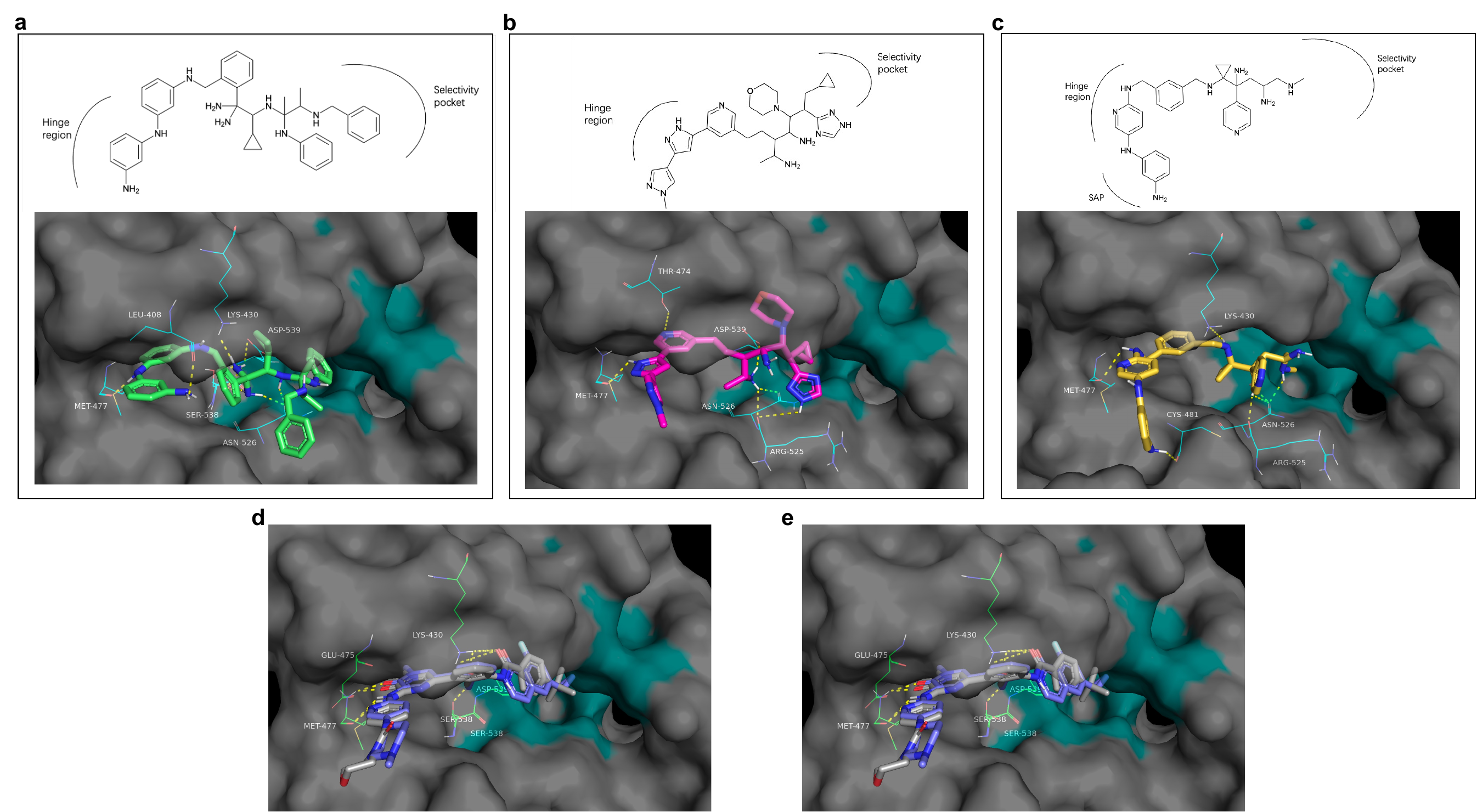}
\caption{Structure-based design of BTK inhibitors using HelixMO. a-c) Examples of generated compounds. Top panel: Structure of a) SE4-244 b) SE3-388 and c) SE3-166. Bottom panel: binding poses of a) SE4-244 b) SE3-388 and c) SE3-166 in the BTK pocket. d-e) binding poses of d) CGI-1746 and e) RN486 in the BTK pocket. a-e) ligands are shown as stick; interacting residues are shown as cyan lines; interactions between the ligand and the residues are shown as yellow dot lines; the selectivity pocket is shown as cyan surface.}
\label{fig:btk}
\end{figure*}

To illustrate the potential of HelixMO in drug design, we take BTK as an example target and attempt to produce feasible BTK inhibitors. BTK is a key player in modulating B cell development and proliferation. Overexpression of this protein has been observed in B cell leukemia and lymphomas \cite{weber2017bruton}. In fact, the use of BTK inhibitors in patients with malignant lymphoma led to positive outcomes, and several BTK reversible/irreversible inhibitors have been approved or are in different stages of clinical trials (Reviewed in \cite{liu2021emerging}). Representative molecules designed by HelixMO are shown in Fig.~\ref{fig:btk}a - Fig.~\ref{fig:btk}c. SE4-244 and SE3-388 adopt similar modes of interaction with reported  BTK inhibitors CGI-1746 and RN486 (Fig.~\ref{fig:btk}d and Fig.~\ref{fig:btk}e) \cite{lou2015structure}. The two generated molecules form hydrogen bonds with MET477 in the hinge region, and the linkers are stabilized by other residues in the pocket, including LYS430 and ASP539. The cyclopropane and phenyl tail point towards the H3 pocket, occupied of which is important for ligand selectivity. Another generated compound SE3-166 (Figure~\ref{fig:btk} c), while exhibiting the above-mentioned interaction with MET477 and LYS430, its amino phenyl group also goes into the solvent-accessible pocket (SAP) and forms a hydrogen bond interaction with Cys481. Cys481 is a key residue in the BTK catalytic domain and is targeted by five approved BTK inhibitors. Ligands interacting with this residue exhibit improved potency and selectivity\cite{liu2021emerging}. Altogether, our proof-of-concept case study demonstrates that HelixMO can generate molecules that interact with target protein in a similar manner as known active molecules.

\end{appendices}

\footnotesize
\bibliographystyle{unsrt}  
\bibliography{main}  

\end{document}